\pdfoutput=1

\documentclass[11pt]{article}

\usepackage[]{acl}

\usepackage{times}
\usepackage{latexsym}
\usepackage{tikz-dependency}
\usepackage{booktabs}
\usepackage{tcolorbox}
\usepackage{tikz,pgfplots}
\usepgfplotslibrary{groupplots}
\usepackage{nicefrac} 
\pgfplotsset{compat=1.13}

\definecolor{g-red}{HTML}{DB4437}
\definecolor{g-blue}{HTML}{4285F4}
\definecolor{g-green}{HTML}{0F9D58}
\definecolor{g-yellow}{HTML}{F4B400}
\definecolor{g-orange}{HTML}{FF9800}
\definecolor{g-grey}{HTML}{9E9E9E}
\definecolor{shannon}{HTML}{304FFE}
\definecolor{uw}{RGB}{138,43,226}
\definecolor{stanford}{RGB}{255,69,0}
\definecolor{const}{RGB}{68, 110, 182}
\definecolor{head}{RGB}{246, 180, 32}
\definecolor{freq}{RGB}{0, 0, 0}
\definecolor{ao}{rgb}{0.0, 0.5, 0.0}
\definecolor{asparagus}{rgb}{0.53, 0.66, 0.42}
\definecolor{amber}{rgb}{1.0, 0.49, 0.0}
\definecolor{alizarin}{rgb}{0.82, 0.1, 0.26}
\definecolor{applegreen}{rgb}{0.55, 0.71, 0.0}
\definecolor{amethyst}{rgb}{0.6, 0.4, 0.8}
\definecolor{auburn}{rgb}{0.43, 0.21, 0.1}

  {\begin{itemize}[topsep=0.5pt, partopsep=0.5pt] %
    \setlength{\itemsep}{0.5pt}%
    \setlength{\parskip}{0.5pt}%
    }%
  {\end{itemize}}

\usepackage{caption}
\usepackage{booktabs}

\usepackage{babel}

\renewenvironment{quote}{%
  \list{}{%
    \leftmargin0.3cm   
    \rightmargin\leftmargin
  }
  \item\relax
}
{\endlist}

\usepackage{comment}
\usepackage{graphicx} 
\usepackage{amssymb}

\usepackage{pifont}
\usepackage[T1]{fontenc}
\usepackage[utf8]{inputenc}

\usepackage{microtype}

\usepackage{amsmath,amsfonts,bm}

\def\eqref#1{equation~\ref{#1}}

\def\1{\bm{1}}

\DeclareMathAlphabet{\mathsfit}{\encodingdefault}{\sfdefault}{m}{sl}
\SetMathAlphabet{\mathsfit}{bold}{\encodingdefault}{\sfdefault}{bx}{n}

\newcommand{\heart}{\text{\small \ding{170}}}

\title{Pushing the Limits of ChatGPT on NLP Tasks}

\author{Xiaofei Sun$^{\spadesuit}$, Linfeng Dong$^{\spadesuit}$, Xiaoya Li$^{\clubsuit}$, Zhen Wan$^{\blacklozenge}$ \\
{\bf Shuhe Wang$^{\clubsuit}$, Tianwei Zhang$^{\blacktriangle}$, 
Jiwei Li$^{\spadesuit}$, 
Fei Cheng$^{\blacklozenge}$}\\
{\bf Lingjuan Lyu$^{\blacktriangledown}$, Fei Wu$^{\spadesuit}$, Guoyin Wang$^{\heart}$}}

\begin{document}

\maketitle
\begin{abstract}
Despite the success of ChatGPT, 
its performances on most NLP tasks
are still well below the supervised baselines. 
In this work, we looked into the causes, and discovered that its subpar performance was caused by the following factors:
(1) token limit in the prompt does not allow for the full utilization of the supervised datasets;
(2) mismatch between the generation nature of ChatGPT and NLP tasks;
(3) intrinsic pitfalls of LLMs models, e.g., hallucination, overly focus on certain keywords, etc.

In this work, we propose 
a collection of general modules to address these issues, in an attempt to push the limits of ChatGPT on NLP tasks. 
Our proposed modules include 
(1) a one-input-multiple-prompts strategy that
employs multiple prompts for one input to accommodate more demonstrations;
(2) using fine-tuned models for better demonstration retrieval;
(3) transforming tasks to formats that are more tailored to the generation nature; 
(4) employing reasoning strategies that are tailored to addressing the task-specific complexity;  
 (5) the self-verification strategy to address the hallucination issue of LLMs; (6) the paraphrase strategy to improve the robustness of  model predictions. 

We conduct experiments on 21 datasets of  10 representative NLP tasks,
including 
question answering, commonsense reasoning, natural language inference, sentiment analysis, named entity recognition, entity-relation extraction, event extraction, dependency parsing, semantic role labeling, and part-of-speech tagging. 
Using the proposed assemble of techniques, 
we are able to significantly boost the performance of ChatGPT on the selected NLP tasks,
achieving performances comparable to or better than supervised baselines, or even existing SOTA performances. 
\footnote{~$^{\spadesuit}$Zhejiang University, $^{\clubsuit}$Shannon.AI, $^{\blacklozenge}$Kyoto University, $^\blacktriangle$Nanyang Technological University, $^\blacktriangledown$Sony AI, $^{\heart}$Amazon}
\footnote{xiaofei\_sun@zju.edu.cn}
\end{abstract}
\section{Introduction}

Despite the success achieved by ChatGPT\footnote{https://openai.com/blog/chatgpt}, its performances on most NLP tasks are still significantly below supervised baselines~\citep{qin2023chatgpt}.
This is due to the following reasons:
(1) 
token limit:
there is a hard token limit (4,096)
for the input to the ChatGPT, which means only a small fraction of the labeled data can be used in the prompt for in-context learning (ICL);
On the contrary, supervised baselines can harness the full labeled dataset;
(2) the mismatch between ChatGPT and many NLP tasks:
ChatGPT is a text generation model, while many NLP tasks cannot be easily formatted as a text generation task, e.g., named entity recognition (NER), dependency parsing, semantic role labeling, etc. 
The adaptation from the original NLP task to a text generation task comes at a heavy cost in performance; 
(3) the reasoning power of ChatGPT has not been fully fulfilled with respect to different tasks, which may require 
different reasoning ability to address the task-specific language complexity; and
(4) the intrinsic pitfalls from ChatGPT itself: ChatGPT severely suffers from the hallucination issue~\citep{ji2023survey}, where in the context of NLP tasks, 
it tends to overconfidently label \textit{null} instances with labels that they don't belong to. 

In this paper, we explore how we can systematically address the aforementioned issues of ChatGPT, in an attempt to push the limit of its performances 
on different NLP tasks.
Strategies we explore include  
(1) the one-input-multiple-prompts strategy that
employs multiple prompts for one input to accommodate more demonstrations;
(2) using fine-tuned models for better demonstration retrieval to make every token in the prompt count;
(3) proper task formalizations to
transform tasks to formats that are more tailored to the generation nature; 
(4) employing reasoning strategies that are tailored to address the task-specific complexity;  
(5) the self-verification strategy to address the hallucination issue of LLMs; and (6) the paraphrase strategy to improve the robustness of  model predictions. 

With the combination of these strategies, we are able to significantly boost the performance of ChatGPT on all selected NLP tasks.
We conduct experiments on 21 datasets of 10 representative NLP tasks,
including question answering, commonsense reasoning, natural language inference, sentiment analysis, named entity recognition, entity-relation extraction, event extraction, 
dependency parsing, semantic role labeling, and part-of-speech tagging. 
Using the proposed assemble of techniques, 
we are able to significantly boost the performance of ChatGPT on the selected NLP tasks,
achieving performances comparable to or better than supervised baselines, or even existing SOTA performances. 

\section{Methodology}
In this section, we detail strategies to address the aforementioned disadvantages of the ChatGPT system  in order below.
\subsection{One-input-multiple-prompts}
For ChatGPT, there is a hard limit of 4,096 token in the input. Therefore, only a very small fraction of 
training examples can be used. 
To address this issue, we propose the {\it one-input-multiple-prompts} strategy. 
Let $N$ denotes the number of prompts for each input.
Each prompt is filled with distinct demonstrations. Demonstrations are retrieved using  random or $k$NN strategies. 
Prompts are fed to ChatGPT separately. 
We thus will get $N$ predictions from ChatGPT.
The final result 
is made via voting among the individual judgments made by ChatGPT for each prompt.
By doing this, we can get around the restriction on tokens, allowing us to take the advantage of more training data.

\subsection{The FT-retrieval strategy}
Another direction to address the limited token issue is to improve the quality of demonstrations to make every token in the prompt count. 
$k$NN-based retrieval is
 based on general sentence-level representations~\citep{gao2021simcse, sun2022sentence, seonwoo2022ranking}
and retrieves similar demonstrations in terms of the general semantic.
They
surely perform better than random retrieval, but come with the key disadvantage: 
they 
do not extract features tailored to the specific task. 
A better alternative is to use the fine-tuned (FT for short) model on the training set as the similarity measurement function. 
Specifically,
we first fine-tune a supervised model (e.g., RoBERTa~\citep{liu2019roberta}) based on the full training set, 
and use representations from the fine-tuned model for $k$NN search.
For sentence-level tasks such as sentiment analysis or natural language inference, we use sentence-level representations from the FT model, 
while for token-level tasks such as entity extraction or event extraction, we use token-level representations for retrieval. 
From a global perspective, the FT-retrieval strategy bridges the gap between ChatGPT and the supervised model:
though ChatGPT cannot fully the training data due to the token limit, we can still setup their connection through the FT retrieval since the latter 
is trained based on the full training data.

\subsection{The reasoning strategy}

\citet{wei2022chain} propose the chain-of-thoughts (COT) strategy to enhance LLMs' reasoning abilities for solving math tasks: COT first generates intermediate rationale explanations and then followed by the task-related decision.  
Prompting LLMs to generate chain-of-thoughts before making decisions can improve LLMs' inference capabilities for a given task, reduce the randomness of model decoding and enhance LLMs' performance on complex reasoning tasks.
However, COT cannot be readily applied in many of NLP scenarios because 
it has the fixed demonstration issue:
for COT,
each test input uses the same set of demonstrations, 
the reasoning of 
which is prepared in advance, and  
 only a small number of (e.g., 8, 16) training instances are selected as demonstrations.
For the $k$NN strategy, which is adopted in most NLP scenarios, 
each instance in the training set has a chance to be selected. Therefore, we need to prepare intermediate reasoning explanations for all training examples.
To address  this issue, we propose to use ChatGPT to generate rationale for all training examples. 
Specifically, 
at the rational-preparing stage,
we first transform each data \textsc{(input, label)} in the training set to  \textsc{(input, rationale explanations, label)} by prompting
ChatGPT
to generate intermediate rationale explanations that support model decisions. 
At test time, we feed the concatenation of the task description, demonstrations that involve rationales, and the test instance to ChatGPT, in which case
ChatGPT should generate a string that includes the reasoning process of ChatGPT, followed by its task-related decision for the input test.

We use the natural language inference task to illustrate the automatic rational generation stage:  
given an example with 
premise: "\textit{A man inspects the uniform of a figure in some East Asian country}"; and hypothesis: "\textit{The man is sleeping}", and the NLI label \textit{contradiction}.
We prompt ChatGPT to generate the rational for the label 
{\it contradiction}
between the premise and the hypothesis, which is as follows: 
\begin{quote}
{\it The hypothesis that the man is sleeping contradicts the premise that the man is inspecting the uniform. This is because the two statements describe mutually exclusive states of being. If the man was sleeping, he would not be able to actively inspect anything, including a uniform. Therefore, the hypothesis does not entail the premise.} 
\end{quote}
The automatically generated rational will be stored and later used at the $k$NN stage.

\begin{figure*}[!t]
 \center
 \begin{minipage}[b]{1.8\columnwidth}
 \vspace{-1.4cm}
\hspace{-0.1cm}\includegraphics[scale=0.23]{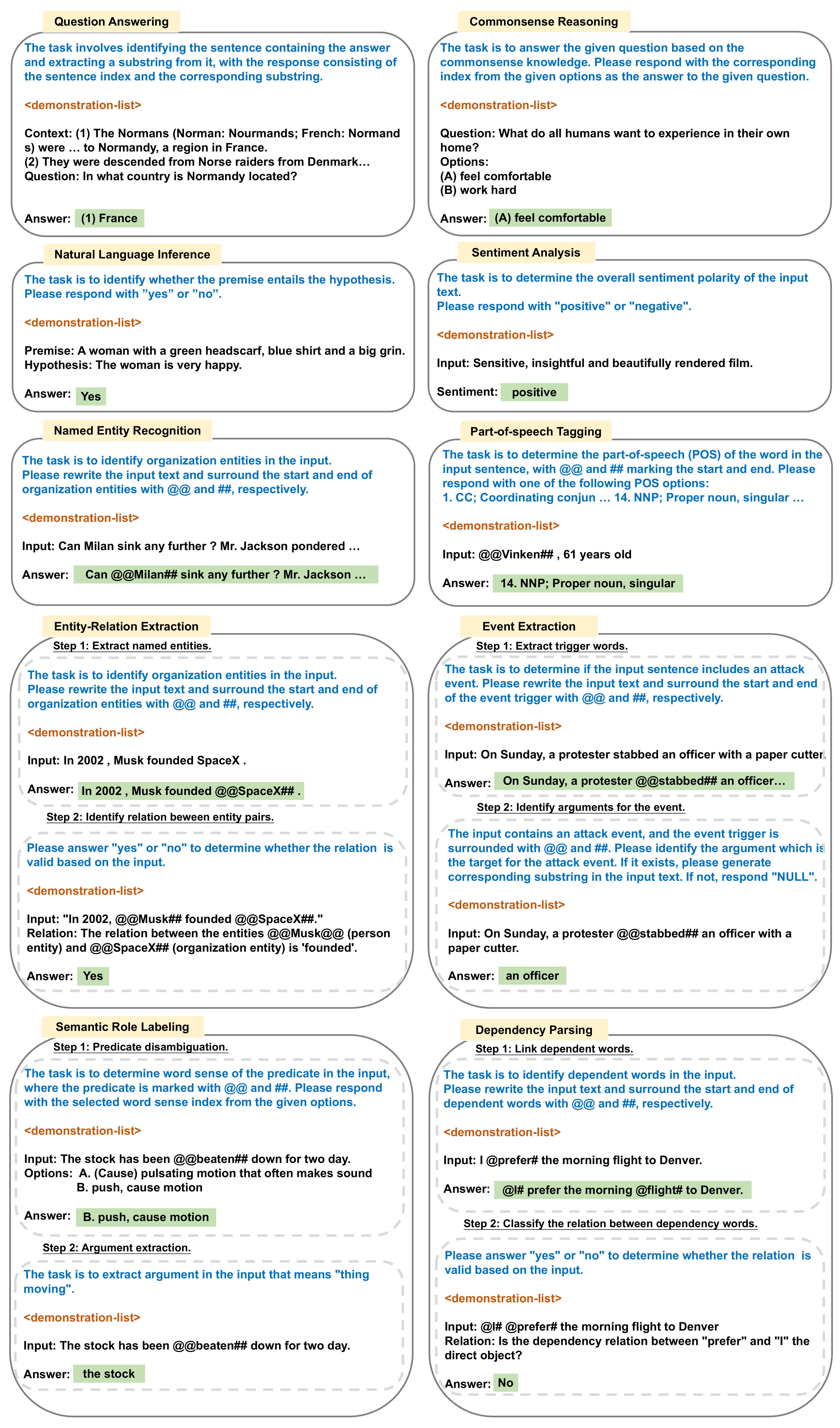}
\caption{Task Formalizations under ChatGPT, including question answering, commonsense reasoning, natural language inference, sentiment analysis, named entity recognition, entity-relation extraction, event extraction, dependency parsing, semantic role labeling, and part-of-speech tagging. }
\label{fig:task_formalization}
 \end{minipage}
 \end{figure*}


\subsection{Proper Task Formalization}
ChatGPT is a text generation model in nature. Unfortunately, 
many NLP tasks cannot be easily formalized as a text generation task (e.g., NER, dependency parsing, etc),
leading to inferior performances.
We propose a few strategies to 
 structure NLP tasks in a generative manner that is more suited to  text generation, which can lead to smaller cost in performance. 
 
The first effective recipe we find effective is to 
 prompt ChatGPT to
copy the input while modifying 
 labeled tokens by surrounding them  with special symbols.  
For example, to extract location (LOC) entities in the input "{\it he lives in Seattle}" in the NER tasks, 
the output from ChatGPT surrounds the LOC entity "{\it Seattle}" with special symbols \#\# and @@,
making the output “{\it he lives in  \#\# Seattle@@}”.
Take the dependency parsing task as another example, 
if we want to extract the dependents of word "{\it like}" in the sentence "{\it I like the movie}", 
the output from ChatGPT should be ”{\it  \#\#I@@ like the  \#\#movie@@}“, marking that ”{\it I}“ and “{\it "movie}”
are extracted words.
This copy-and-modify approach not only preserves the continuity of the output, but also simplifies the process of connecting the output to the extracted tokens, resulting in superior results compared to other methods. 

The second recipe is to transform the N-class multi-class classification task to N {\bf binary classification} tasks.
For example, for
the task of natural language inference, 
which is  an N-class multi-classification task, the most straightforward strategy is to transform the task to a multiple-choice problem, e.g.,
{\it decide the  logical relation between sentenceA and sentenceB: (a) entailment (b) contradiction (c) neutral}. 
Instead of using the multiple-choice strategy, 
the N-binary classification strategy
transforms the task to N binary classification tasks: 
{\it does sentenceA entails sentenceB, yes or no?};
{\it does sentenceA contradicts with sentenceB, yes or no?}. 
The intuition is that, for each class, we are able to show more illustrations with respect to that class with the binary-transformation strategy.
The 
N-class multi-class classification comes at the cost of significantly more queries to ChatGPT. 
When there is an extremely large number of categories, such as 45 for Part-of-Speech tagging, using the binary-transformation method can be prohibitively expensive. 
In these cases, we will go back to the 
multi-choice strategy.

\subsection{The Self-Verification Strategy}
ChatGPT suffers from the hallucination issue~\citep{ji2023survey}, which generates false positive predictions with high confidence.
Using the named entity recognition task as an example, the hallucination issue refers to ChatGPT extracting entities from sentences that do not contain any entities.

We propose the self-verification strategy (SV for short) to address the above issue. 
Specifically, the self-verification strategy is to conduct a new round of validation after obtaining generated results of ChatGPT for a NLP task.
After obtaining the generated task results from ChatGPT, we concatenate the task description with the generated result and ask ChatGPT answer whether the generated result is correct or not.
ChatGPT will generate a "\textit{yes}" or "\textit{no}" to determine whether the generated results are reasonable for the original task.
Therefore, the hallucination issue can be addressed by using the self-verification strategy to validate the generated results of ChatGPT and correcting erroneous predictions.

Let's take the named entity recognition task as an example to illustrate. Given the input "\textit{Hunan Office in Beijing}". 
ChatGPT has completed the first step of extracting the location (LOC) entity and identified "Hunan" as a LOC entity.
We employ the self-verification strategy to validate the LOC result obtained in the first step. We prompt ChatGPT: 
\begin{quote}
\textit{INPUT: \@Hunan\# Office in Beijing 
\newline 
Based on the context, is the labeled 'Hunan' in the INPUT a location entity?}
\end{quote}
ChatGPT should generate "\textit{no}" indicating that "\textit{Hunan}" is not a location entity.
Afterward, we remove "\textit{Hunan}" from the list of LOC entities predicted by ChatGPT in the first step.

\subsection{The Paraphrase Strategy}

ChatGPT often faces the issue that predictions are dominated by surface words. 
This is due to the limited demonstrations in prompts, and ChatGPT 
sometimes ignores the context information and is mislead by high-frequency words or words with strong task-related features, resulting in inferior performances. 
Using the question-answering task as an example for an illustration:  given the context "\textit{The news agency reports that the goverment ...}", and the question "\textit{What is the topic of the input text?}",
ChatGPT is dominated by the phrase "\textit{The news agency}" and generates "\textit{news}" as the answer to the given question. 

To address the surface word domination issue, we propose the paraphrase strategy. Specifically, we use ChatGPT to paraphrase the given text and get multiple 
versions of the input with the same meaning but in different expressions. Next, we use paraphrases as the input to ChatGPT, one at a time. 
Since we will get multiple decisions with different paraphrases, a voting strategy is employed to obtain the final decision. 
The fact that words in the original test example are likely to be replaced with different expressions makes ChatGPT less immune to specific tokens, but focus on the overall semantics, we are able to get more robust results.

It is worth noting that the paraphrase strategy can only be applicable to sentence-level tasks, but not token-level tasks (e.g., NER, POS). 
This is because words in the generated paraphrases usually cannot be accurately aligned back to the original input.

\begin{table}[t]
\centering
\small
\scalebox{0.85}{
\begin{tabular}{lccc}
\toprule
& {\bf SQuADv2}& {\bf TQA} & {\bf MRQA-OOD} \\
{\bf Model}& {\bf (EM)}  & {\bf (EM)} &{\bf (F1)} \\
\midrule
RoBERTa-Large & 86.8 & {\bf 81.1} & 72.4 \\\midrule
\textit{\bf ChatGPT}~\textit{(few-shot)}  &  &  &   \\
+Random demo & 70.1 & 70.9 & 63.1  \\
+SimCSE $k$NN & 73.5 & 72.5 & 65.7  \\
+FT $k$NN & 78.9 & 75.8 & 68.3  \\
+FT $k$NN+Multi& 83.6 & 78.0 & 71.0  \\
+FT $k$NN+Multi+Reason& 87.2 & 79.3 & 75.6  \\
+FT $k$NN+Multi+Reason+SV& {\bf 88.2} & 80.8 & {\bf 76.1}  \\
\bottomrule
\end{tabular}
}
\caption{Experimental results for the question answering task. We abbreviate the self-verification strategy as SV.}
\label{exp:qa}
\end{table}
\section{Tasks and Performances}

\subsection{Question Answering}
\subsubsection{Task Description}
Question answering (QA)~\citep{ Seo2016BidirectionalAF,  Xiong2016DynamicCN, Wang2017GatedSN, 
devlin2018bert, Peters2018DeepCW, Yu2018QANetCL, Li2019DiceLF,  yang2019xlnet} is a task that generates an answer to a given natural language question.
QA under the framework of machine reading comprehension, where the answer is a substring of the given context, is to extract a substring from the context as the answer to the question.
The task is normally formalized as a multi-class classification problem, which determines whether the current word is the start (i.e., S label), end (i.e., E label), or not part of the answer (i.e., O label). 
If there is no start or end in the label sequence, the question is not answerable according to the given text.

\subsubsection{Formalization under ChatGPT}
Under ChatGPT, QA is formalized as a text generation task, where ChatGPT is directly prompted to generate the answer text for the given context and the question.

We first split the given context into individual sentences and assign them a sentence index based on their position in the context. Then we concatenate the modified context and the given question to elicit a response from ChatGPT. 
The generated text string from ChatGPT should consist of two components: (1) the index of the sentence of which the answer is a substring; and (2) the substring that answers the question.
For example, 
with the prompt to ChatGPT being: 
\begin{quote}
\textit{Context: (1) The capital of Japan is Tokyo. (2) The capital of China is Beijing. (3) The capital of South Korea is Seoul. \newline Question: What is the capital of South Korea?}
\end{quote}
 ChatGPT should output 
 \begin{quote}
 \textit{(3) Seoul}
 \end{quote}
 where "\textit{(3)}" denotes the index of the sentence where the answer is located and "\textit{Seoul}" represents the answer.
This strategy provides the model with further guidance by first predicting the index of the sentence within the context, and then deciding which substring in that sentence should be used as the answer,
a strategy akin to multi-task learning. 
 Examples of the task formalization are shown in Figure~\ref{fig:task_formalization}

\subsubsection{Datasets and Results} 
We evaluate on three
widely-used QA benchmarks, i.e., SQuADv2~\citep{rajpurkar2018know}, TQA~\citep{joshi2017triviaqa}, and MRQA-OOD.
More details of the datasets can be found in Appendix~\ref{appendix:dataset_qa}. 

Experimental results are shown in Table~\ref{exp:qa}.  
As shown in Table~\ref{exp:qa}, the
SimCSE-$k$NN retriever outperforms the random retriever, which demonstrates the importance of selecting semantically similar examples as demonstrations for QA. 
FT-$k$NN introduces a significant performance boost over SimCSE-$k$NN.
This indicates that using the FT model, which is fine-tuned on the given training set,  retrieves similar examples with respect to the specific task, and can help improve ChatGPT's performances.

Regarding prompting strategies, we  observe that using the multiple-prompt strategy obtains a significant performance boost across three QA datasets, i.e., +4.7, +2.2, +2.7, respectively on SQuAD V2.0, TQA, and MRQA-OOD datasets. This demonstrates that the multiple-prompt strategy effectively addresses the input token limit issue and allows ChatGPT to take advantage of more annotated examples.

Besides, we find that the rational-based prompting strategy can further boost the performances, +1.8 on SQuADv2, +1.1 on TQA, and +0.8 on MRQA-OOD. This phenomenon is in line with our expectation that intermediate  rationales enhances models' reasoning abilities. 
The proposed self-verification strategy introduces further performance boosts, i.e., +1.0, +1.5, and + 1.5 on SQuADv2, TQA, and MRQA-OOD, respectively.

In general, with the proposed series of strategies, ChatGPT is  able  to achieve comparable performances to the supervised baselines across the three datasets. 
In the out-of-domain setting of MRQA, ChatGPT significantly outperforms the supervised RoBERTa model by +3.7, which indicates  
the significantly better domain-adaptable ability of ChatGPT. 
For TQA and SQuADv2,
ChatGPT performs slightly worse (-0.3)  on the former and slightly better (+1.4) on the latter.

\begin{table}[t]
\center
\small
\scalebox{0.9}{
\begin{tabular}{lcc}\toprule 
& {\bf CSQA} &  {\bf StrategyQA}  \\
{\bf Models} & {\bf (ACC)} &  {\bf (ACC)}  \\\midrule
RoBERTa-Large & {\bf 79.2} & {\bf 72.0} \\\midrule
\textit{\bf ChatGPT}~\textit{(few-shot)}  &   \\
+Random demo & 74.8 &	59.5   \\
+SimCSE $k$NN &  74.7 & 59.6 \\
+FT $k$NN & 76.6& 65.4\\
+FT $k$NN+Multi& 78.0 & 67.8 \\
+FT $k$NN+Multi+Reason & 78.2& 69.4 \\
+FT $k$NN+Multi+Reason+SV& 79.0& 69.9\\
\bottomrule
\end{tabular}
}
\caption{Experimental results on commonsense reasoning datasets. We abbreviate the self-verification strategy as SV.}
\label{exp:commonsense}
\end{table} 


\subsection{Commonsense Reasoning}
\subsubsection{Task Description}
Commonsense reasoning~\citep{bailey2015winograd, Trinh2018ASM, Rajani2019ExplainYL, Lin2019KagNetKG, Liu2020KGBARTKG, sap2020introductory, Klein2020ContrastiveSL}  is a task that uses human consensus and logical inference abilities to generate an answer to a given natural language question. 
Current commonsense reasoning datasets are in the format of multi-choice questions, which 
offer a selection of answers to choose from with one being the right one.


The commonsense reasoning task is normally formalized as a binary classification task.
Models encode the given question and several answer options, assigning the label "I" to the correct answer and the label "O" to other options.

\subsubsection{Formalization under ChatGPT}

Under ChatGPT, commonsense reasoning is formalized as a text completion task to copy the right answer from the given multi-choice options. 
For example, the question is "\textit{Where on a river can you hold a cup upright to catch water on a sunny day?}", the answer choices are
"\textit{(A) waterfall; (B) bridge; (C) valley; (D) pebble}"
and the prompt to ChatGPT is:
\begin{quote}
\textit{Please select the answer to the question from several options. \newline Question: Where on a river can you hold a cup upright to catch water on a sunny day? \newline Options: (A) waterfall; (B) bridge; (C) valley; (D) pebble}. 
\end{quote}
The ChatGPT should generate "\textit{(D) pebble}" which denotes that the \textit{pebble} is the answer to the given question. 
Examples of the task formalization are shown in Figure~\ref{fig:task_formalization}.

\subsubsection{Datasets and Results}

We conduct experiments on two widely-used benchmarks: CSQA~\citep{talmor2018commonsenseqa} and StrategyQA~\citep{geva2021did}.
More details of the datasets can be found in Appendix~\ref{appendix:dataset_commonsense}.

As can be seen from Table 2, the performance of using SimCSE to retrieve $k$NN strategy is comparable to the randomly selecting examples strategy, performing slightly better on the CSQA dataset but lagging slightly behind on the StrategyQA dataset. Using the fine-tuned model to retrieve $k$NN introduces a huge performance boost compared with the SimCSE and the random selection strategies, i.e., +1.9 and +5.8 on the CSQA and StrategyQA dataset, respectively. 

For the multiple-prompt strategy, we  observe that it gains significant performance boosts on two commonsense reasoning benchmarks: +1.4 on the CSQA and +2.4 on the StrategyQA datasets. 
Additionally, 
we find that the reasoning strategy introduces performance improvements across the two benchmarks: a huge performance boost (i.e., +1.6) on the StrategyQA datasets; a slight performance improvement on the CSQA dataset. 
For the self-verification strategy, we observe consistent performance boosts on the CSQA and StrategyQA datasets, i.e., +0.8 and +0.5 increases, respectively. 

Overall, ChatGPT underperforms the supervised baseline on the commonsense reasoning benchmarks. Specifically, ChatGPT performs slightly worse than the RoBERTa on the CSQA, and is lower than the RoBERTa by -2.1 on the StrategyQA dataset. 
The reason for this phenomenon is that ChatGPT's performance on the commonsense reasoning task is  severely limited by the hallucination issue, which can be partially addressed by the proposed self-verification strategy, but not completely. 

\begin{figure*}[!t]
 \center
 \begin{minipage}[b]{1.8\columnwidth}
 \vspace{-1.6cm}
\hspace{-0.5cm}\includegraphics[scale=0.42]{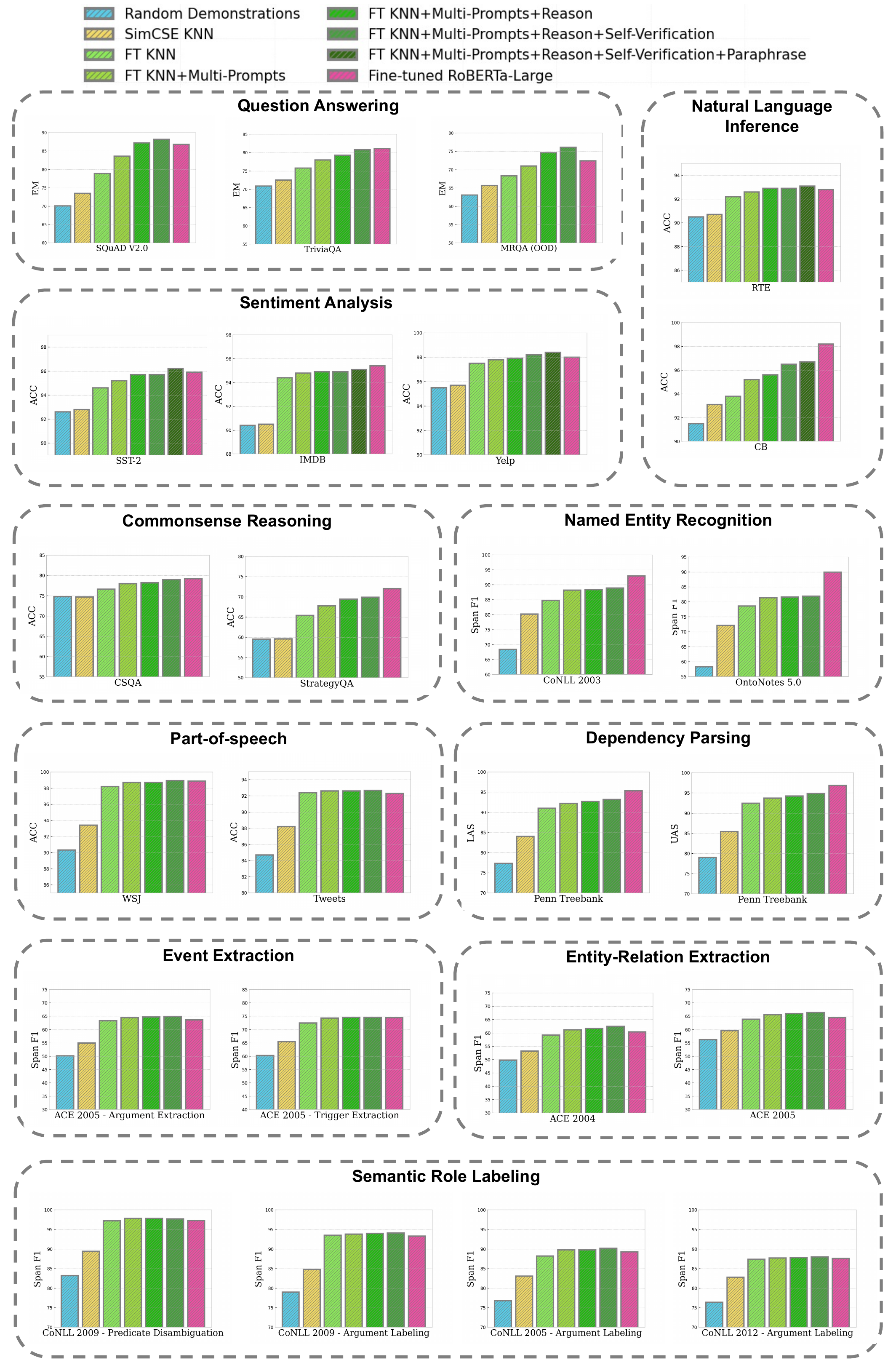}
\caption{Comparisons of experiment results on ten NLP downstream tasks.}
\label{fig:exp_result}
\end{minipage}
\end{figure*}

\begin{table}[t]
\center
\small
\scalebox{0.9}{
\begin{tabular}{lcc}\toprule 
 & {\bf RTE} &  {\bf CB} \\
{\bf Models} & {\bf (ACC)} &  {\bf (ACC)} \\\midrule
RoBERTa-Large  & 92.8 & {\bf 98.2}\\\midrule
\textit{\bf ChatGPT}~\textit{(few-shot)} &  & \\
+Random demo  & 90.5 & 90.5 \\
+SimCSE $k$NN  & 90.7 & 90.4  \\
+FT $k$NN & 92.2  & 93.8 \\
+FT $k$NN+Multi & 92.6 & 95.2\\
+FT $k$NN+Multi+Reason & 92.9 & 95.6 \\
+FT $k$NN+Multi+Reason+SV &  92.9 & 96.5 \\
+FT $k$NN+Multi+Reason+SV+Paraphrase & {\bf 93.1} & 96.7\\
\bottomrule
\end{tabular}
}
\caption{Experiment results on natural language inference benchmarks. We abbreviate the self-verification strategy as SV.}
\label{exp:inference}
\end{table} 

\subsection{Natural language inference}
\subsubsection{Task Description}
Natural language inference (NLI)~\citep{Wang2015LearningNL, Mou2015NaturalLI, Liu2016LearningNL, Parikh2016ADA, Chen2016EnhancedLF, Conneau2017SupervisedLO, Chen2017NeuralNL,  camburu2018snli,  McCoy2019RightFT,  Yan2021ConSERTAC} is a task that aims to determine whether the given hypothesis can be logically inferred from the given premise: the premise entails the hypothesis (i.e., entailment); the hypothesis and the premise are contradictions (i.e., contradiction); and the hypothesis is undetermined given the premise (i.e., neural). The NLI task is normally formalized as a three-class classification problem. Models first encode the given premise and the given hypothesis into a feature vector. Then the combined feature vector is passed to a classifier.

\subsubsection{Formalization under ChatGPT}
Under ChatGPT, NLI is formalized as prompting ChatGPT to generate {\it yes/no} with respect to each logical relation (e.g., entailment), given the premise and the hypothesis. If the response is \textit{yes}, it denotes that the relation holds between the premise and the hypothesis. Since there are three candidate logical relations, the prompting process should be repeated three times.   
For example, given the premise "\textit{Pibul Songgramwas the pro-Japanese military dictator of Thailand during World War 2.}", the hypothesis "\textit{Pibul was the dictator of Thailand.}", and the prompts to ChatGPT is 
\begin{quote}
\textit{Does the premise entail the hypothesis? Please answer yes or no. \newline 
Premise: <premise> \newline 
Hypothesis: <hypothesis>}
\end{quote}
where \textit{<premise>} and \textit{<hypothesis>} should be replaced by the given premise and hypothesis, respectively. ChatGPT should output "\textit{no}" ideally in this case, which denotes that the premise does not entail the hypothesis.  Subsequently, the same procedures are used to verify the contradiction and neural relations. Examples of the task formalization are shown in Figure ~\ref{fig:task_formalization}.

\subsubsection{Datasets and Results}
We conduct experiments on two benchmarks: RTE, CommitmentBank~(CB)~\citep{de2019commitmentbank}. 
More details of the datasets can be found in Appendix~\ref{appendix:dataset_nli}. 
We use accuracy in three-class (e.g., entail, contrast, neural) as the evaluation metric.

Experimental results are shown in Table~\ref{exp:inference}.
As can be seen, 
ChatGPT with SimCSE retriever performs much better than the random selection strategy on RTE and the CB datasets, +0.2, and +1.6, respectively. 
Notably, using the fine-tuned model as a demonstration retriever significantly improves the performance on the two datasets, +1.5 on RTE and +0.7 on CB. This suggests that retrieving demonstrations with a fine-tuned model can help ChatGPT
retrieve better demonstrations with respect to the task, leading to better performances.

For the multiple-prompt strategy, we  observe that the multiple-prompt strategy gains significant performance boosts on two natural language inference benchmarks: +0.4 on the RTE and +1.4 on the CB datasets. 
This phenomenon indicates that the multiple-prompt strategy can effectively address the input token limit issue, leading to better performances. 
Besides, we observe that incorporating reasoning explanations as part of the prompt yields consistent performance improvements across both datasets, +0.3 on RTE and +0.4 on CB. This phenomenon aligns with our expectations that reasoning explanations can improve the inference capability of ChatGPT, and natural language inference is a task that requires reasoning capability,
the performance of which will be boosted with the increasing inference capability. 

For the self-verification strategy, where ChatGPT's wrong predictions are corrected in the subsequent verification step, we are able to achieve a performance improvement +0.9 on the CB dataset, although there is no performance change on the RTE dataset.
For the self-verification strategy, where ChatGPT's wrong predictions are corrected in the subsequent verification step, we are able to achieve consistent performance improvement across both datasets, +6.4 on RTE and +0.9 on CB.

Regarding the paraphrase strategy, where the input text is paraphrased into another text string with the same meaning, we are able to achieve consistent performance improvement across both datasets. 

After using the proposed prompt strategies in the paper, ChatGPT outperforms the vanilla setting (i.e., +0.3 on RTE and +5.2 on CB), and is comparable to the supervised baseline. Specifically, ChatGPT is higher than the RoBERTa model on the RTE dataset by +0.3, and has a lower score than the RoBERTa model on the CB dataset. The reason for this phenomenon is that the natural language inference task requires natural language abilities such as negation, intensification, logical reasoning, and coreference resolution in order to complete the task.

\begin{table}[t]
\center
\small
\scalebox{0.82}{
\begin{tabular}{lccc}\toprule 
 & {\bf SST-2} &  {\bf IMDB} & {\bf Yelp} \\
{\bf Models} & {\bf (ACC)} &  {\bf (ACC)} & {\bf (ACC)} \\\midrule
RoBERTa-Large & 95.9 & {\bf 95.4} & 98.0 \\\midrule
\textit{\bf ChatGPT}~\textit{(few-shot)} &  &  &  \\
+Random demo & 92.6 & 90.4 & 95.5\\
+SimCSE $k$NN   & 92.8 & 90.5 & 95.7\\
+FT $k$NN  & 94.6& 94.4& 97.5 \\
+FT $k$NN+Multi & 95.2& 94.8& 97.8 \\
+FT $k$NN+Multi+Reason & 95.7& 94.9& 97.9 \\
+FT $k$NN+Multi+Reason+SV & 95.7& 94.9& 98.2  \\
+FT $k$NN+Multi+Reason+SV+Paraphrase & {\bf 96.2}& 95.1& {\bf 98.4} \\
\bottomrule
\end{tabular}
}
\caption{Experimental results for the sentiment analysis task. We abbreviate the self-verification strategy as SV.}
\label{exp:sentiment}
\end{table} 

\subsection{Sentiment Analysis}
\subsubsection{Task Description}
Sentiment analysis~\citep{Wilson2005RecognizingCP, Pang2008OpinionMA, Maas2011LearningWV, Taboada2011LexiconBasedMF, Bakshi2016OpinionMA, devlin2018bert,  Xue2018AspectBS,  Ma2018SenticLA, yang2019xlnet,  Basiri2021ABCDMAA} is a task to determine the sentimental polarity (e.g., positive, negative) of a given text. 
The task is normally formalized as a binary or multi-class classification problem, which assigns a sentiment class label to the given text.

\subsubsection{Formalization under ChatGPT}
Under ChatGPT, the task of sentiment analysis can be formalized as 
prompting ChatGPT to
generate sentiment-indicative text given the input (e.g., {\it decide the sentiment of the following text}).
The generated sentiment-indicative text contains sentiment keyword (e.g., positive, negative, etc)
and will be latter mapped to a sentiment label.
Examples of the task formalization are shown in Figure~\ref{fig:task_formalization}.

\subsubsection{Datasets and Results}
We conduct experiments on three benchmarks: SST-2, IMDb, and Yelp. 
More details of the datasets can be found in Appendix~\ref{appendix:dataset_text}. 
We use accuracy as the evaluation metric. 

Experimental results are shown in Table~\ref{exp:sentiment}. 
As can be seen from Table~\ref{exp:sentiment}, using SimCSE to retrieve $k$NN as demonstrations outperforms randomly selected demonstrations across three sentiment analysis datasets. 
We also observe that using the fine-tuned model to retrieve $k$NN gains significant performance boosts 
compared to the SimCSE model, +3.8, +3.9, +1.8 on SST-2, IMDb, and Yelp dataset, respectively. 
The multiple-prompt strategy also yields performance boosts on sentiment analysis datasets. 

We also find that the reasoning strategy can further introduce performance improvements. 
Besides, the proposed 
self-verification (SV) strategy 
yields minor performance improvement only on the Yelp dataset (+0.3).
The explanation is that the performance without SV is already high enough that adding SV provides only a marginal improvement.


By combing the proposed prompt strategies, ChatGPT's performance on SST-2, IMDb, and Yelp are comparable to the supervised baseline, bridging the gap between the supervised baseline and vanilla ChatGPT. Specifically, SST-2 is slightly higher than the supervised baseline (+0.3), while slightly lower than IMDb and Yelp to the supervised baseline. 

\begin{table}[t]
\centering
\small
\scalebox{0.9}{
\begin{tabular}{lcc}
\toprule
 &  \multicolumn{1}{c}{\bf CoNLL 2003}& \multicolumn{1}{c}{\bf OntoNotes 5.0}  \\
{\bf Model}& {\bf  (Span-F1)} &  {\bf  (Span-F1)} \\
\midrule
RoBERTa-Large  & {\bf 93.0}  & {\bf 89.9}  \\\midrule
\textit{\bf ChatGPT}~\textit{(few-shot)} &  &  \\
+Random demo  & 68.4 & 58.3 \\
+SimCSE $k$NN & 80.2 & 72.1  \\
+FT $k$NN  & 84.8 & 78.6  \\
+FT $k$NN+Multi & 88.2 & 81.4  \\
+FT $k$NN+Multi+Reason & 88.4 & 81.6   \\
+FT $k$NN+Multi+Reason+SV & 88.9 & 81.9 \\
\bottomrule
\end{tabular}
}
\caption{Experimental results on the named entity recognition benchmarks. We abbreviate the self-verification strategy as SV.}
\label{exp:ner}
\vskip -0.15in
\end{table}

\subsection{Named entity recognition}
\label{content:ner}
\subsubsection{Task Description}
Named entity recognition (NER)~\citep{Chiu2015NamedER, Lample2016NeuralAF, ma2016end, Ju2018ANL, Li2019AUM, Zhang2018ChineseNU, Ju2018ANL, Shang2018LearningNE, Yan2019TENERAT, Luo2019HierarchicalCR,  Liu2019TowardsIN, wang2022k, wang2022gnn, wang2023gpt} is a task that extracts named entities of pre-defined categories (e.g., location, organization, etc.) from a given text.
The task is normally formalized as a sequence labeling problem, which 
assigns a class label (e.g., O, B-ORG, M-ORG, E-ORG) to each token within a sequence of tokens. 

\subsubsection{Formalization under ChatGPT}
Under ChatGPT, NER is formalized as a text generation task, 
where 
given an input text (e.g., "{\it He lives in Chicago}"),
and a certain entity type (e.g., {\it location}),
we prompt ChatGPT to 
surround 
entities 
belonging to the entity type in the original sequence with special symbols: 
\begin{quote}
\textit{He lives in \#\# Chicago @@}, where
 \#\# and @@ denote the start and end 
of a named entity.
\end{quote}
If there is no location entity in the input, ChatGPT just copies the original input as the output. 
Suppose that there are $N$ types of entities,  
 for each
input sentence, we need to construct the prompt N times,
each of which corresponds to each entity type;
This strategy was  adopted in \citet{wang2023gpt}.
Examples of the task formalization are shown in Figure~\ref{fig:task_formalization}. 

\subsubsection{Datasets and Results}
We conduct experiments on two English benchmarks: CoNLL2003~\citep{sang2003introduction} and OntoNotes5.0~\citep{pradhan2013towards}. 
More details of the datasets can be found in Appendix~\ref{appendix:dataset_ner}. 
We use span-level F1 score as the evaluation metric. 
Experimental results are shown in Table~\ref{exp:ner}.

As can be seen from Table~\ref{exp:ner}, using SimCSE to retrieve $k$NN as demonstrations outperforms using randomly selected demonstrations. 
Further, 
using the fine-tuned model to retrieve $k$NN as demonstrations introduce a huge performance boost over the SimCSE retriever, +4.6 on the CoNLL2003 dataset, +6.5 on the OntoNotes5.0 dataset. The observed phenomenon suggests that demonstrations from the fine-tuned model offer ChatGPT more task-specific features, while demonstrations from the SimCSE model encompass more general-purpose information that is discrepant with the NER task.

We observe that the multiple-prompt strategy introduces large performance gains across two NER datasets, i.e., +3.4 on CoNLL2003, and +2.8 on OntoNotes5.0. 
Further, 
we observe that the reasoning strategy introduces consistently minor performance gains across two datasets. This suggests that the reasoning strategy can improve ChatGPT's inference capacity on NER to a limited extent. 
Besides, we observe that consistent performance boosts are  introduced by the self-verification strategy, i.e., +0.5 on CoNLL2003, and +0.3 on OntoNotes5.0.

Overall, after applying the proposed strategies, we are able to narrow the performance gap between the vanilla ChatGPT and the supervised baseline, though still underperform the supervised baseline. 
Specifically, for the CoNLL2003 dataset, ChatGPT is worse than the RoBERTa by -4.1, and outperforms the vanilla ChatGPT by +20.5; for the OntoNotes5.0 dataset, ChatGPT is worse than the RoBERTa by -8.0, and outperforms the vanilla ChatGPT by +23.6. The reason for the large performance gap between the proposed strategies and the RoBERTa 
is because of the gap between the NER task and the 
 generative nature of ChatGPT.

\subsection{Entity-Relation Extraction}

\subsubsection{Task Description}
Entity-relation extraction~\citep{Mintz2009DistantSF, Li2014IncrementalJE, Miwa2016EndtoEndRE, li2019entity, wan2023gpt} is a task that aims to extract named entities in a given text, and identify relations between the extracted entity pairs. 
The entity-relation extraction task is normally formalized as a two-stage problem: 
(1) assigning an entity label (e.g., O, B-ORG, M-ORG, B-PER) to each word to extract named entities, which can be formalized as a sequence-labeling task;
(2) assigning a relation label  (e.g., founded, give-birth-to) to the extracted entity pairs, which can be formalized as a multi-class classification task.

\subsubsection{Formalization under ChatGPT}
Under ChatGPT, the entity-relation extraction is formalized as a two-step text completion task. \newline
{\bf Step 1}: ChatGPT extracts named entities with respect to a certain type (e.g., location) by rewriting the input sentence and surrounding the entity with special tokens. 
For example,
given the input sentence "\textit{In 2002, Musk founded SpaceX}", we would like to extract entities with respect to the \textit{organization} type.  
The input to ChatGPT is:
\begin{quote}
\textit{Please mark the start and end of ORG entities in the INPUT with @ and \#, respectively. \newline INPUT: In 2002, Musk founded SpaceX. } 
\end{quote}
and ChatGPT should output "\textit{In 2002, Musk founded @SpaceX\#,}" where @ and \# are the starting and ending boundaries of ORG entities and the substring "\textit{SpaceX}" is an ORG entity. If there is no ORG entity in the given text, ChatGPT should copy the input. Suppose the number of NER categories is $N$, the above prompt should be repeated $N$ time for one input. 
The process here is similar to that of NER in Section~\ref{content:ner}. \newline
{\bf Step 2}: Prompt ChatGPT to output a yes-or-no decision to determine whether a certain relation holds between two specified entities.
In the example above,  we have already identified 
the \textit{person}  entity "\textit{Musk}" and the \textit{organization} entity "\textit{SpaceX}" at stage-1, 
the second step involves determining the relation between them. 
Assuming that there are $M$ possible relationships between entities, we ask ChatGPT each relation at a time, 
e.g., regarding the relation type {\it founded},
the input to ChatGPT is:
\begin{quote}
\textit{Please determine whether the relationship between the entities Musk (person) and SpaceX (organization) in the input sentence is 'founded.' Please answer with Yes/No.\newline Input: In 2002, \@Musk\# founded \@SpaceX\#. }
\end{quote}
In this case, 
ChatGPT should generate "\textit{Yes}", indicating that Musk founded SpaceX. 
\subsubsection{Datasets and Results}
We conduct experiments on two widely-used English benchmarks: ACE2004 and ACE2005. 
More details of the datasets can be found in Appendix~\ref{appendix:dataset_re}.

Experimental results are shown in Table~\ref{exp:event}. As can be seen, the SimCSE retriever 
achieves higher performances than the random retriever, and significant performance boost
is introduced by the SimCSE retriever, +6.0 and +4.3 performance boosts in ACE2004 and ACE2005, respectively. 
This shows that the fine-tuned model is more suitable for the entity-relation task, as it is able to retrieve relevant examples pertinent to the task. 

Regarding the multiple-prompt strategy, 
we notice that it introduces a significant improvement on ACE2004 and ACE2005, +2.0, and +1.7, respectively.  
This phenomenon indicates that the multiple-prompt strategy, which 
expands the demonstration set, provides more supervised signals for the task, leading to performance boosts.  
We also observe consistent performance improvements across the datasets when using the reasoning strategy, +0.8, and +0.4 on ACE2004 and ACE2005, respectively. 
Additionally,  significant performance improvements are introduced by
 the self-verification strategy for two entity relation extraction datasets. 

Overall, after using the proposed prompt strategies, ChatGPT's performances  outperform the supervised baseline on the entity-relation extraction datasets, +2.1 on ACE2004 and +1.5 on ACE2005, bridging the huge gap between vanilla ChatGPT and RoBERTa. 

\begin{table}[t]
\centering
\small
\scalebox{0.9}{
\begin{tabular}{lcc}
\toprule
 &  \multicolumn{1}{c}{\bf Trigger}& \multicolumn{1}{c}{\bf Argument} \\
{\bf Model} &{\bf (Span-F1)}  &{\bf (Span-F1)}   \\
\midrule
RoBERTa-Large & 74.5 & 63.6   \\\midrule
\textit{\bf ChatGPT}~\textit{(few-shot)} &  &  \\
+Random demo  & 60.3& 50.2  \\
+SimCSE $k$NN & 65.5& 55.0  \\
+FT $k$NN  & 72.5& 63.3 \\
+FT $k$NN+Multi & 74.3& 64.5  \\
+FT $k$NN+Multi+Reason & 74.6& 64.7  \\
+FT $k$NN+Multi+Reason+SV & {\bf 74.6}& {\bf 64.9}   \\

\bottomrule
\end{tabular}
}
\caption{Experimental results for the event extraction task. We abbreviate the self-verification strategy as SV.}
\label{exp:event}
\vskip -0.15in
\end{table}

\begin{table}[t]
\centering
\small
\scalebox{0.9}{
\begin{tabular}{lcc}
\toprule
 &  \multicolumn{1}{c}{\bf Trigger}& \multicolumn{1}{c}{\bf Argument} \\
{\bf Model} &{\bf (Span-F1)}  &{\bf (Span-F1)}   \\
\midrule
RoBERTa-Large & 74.5 & 63.6   \\\midrule
\textit{\bf ChatGPT}~\textit{(few-shot)} &  &  \\
+Random demo  & 60.3& 50.2  \\
+SimCSE $k$NN & 65.5& 55.0  \\
+FT $k$NN  & 72.5& 63.3 \\
+FT $k$NN+Multi & 74.3& 64.5  \\
+FT $k$NN+Multi+Reason & 74.6& 64.7  \\
+FT $k$NN+Multi+Reason+SV & {\bf 74.6}& {\bf 64.9}   \\

\bottomrule
\end{tabular}
}
\caption{Experimental results for the event extraction task. We abbreviate the self-verification strategy as SV.}
\label{exp:event}
\vskip -0.15in
\end{table}

\subsection{Event extraction}
\subsubsection{Task Description}
Event extraction (EE)~\citep{ahn2006stages, li2013joint, chen2015event, Nguyen2015EventDA, nguyen2016joint,  Nguyen2016ModelingSF, liu2018event, Nguyen2018GraphCN, Liu2018EventDV, Wang2021CLEVECP} is a task that aims to identify the event type and extract information (i.e., trigger, arguments) of an identified event in the given text. 
The task is normally formalized as a two-step classification problem.
The first step is often formalized under a sequence labeling framework, which assigns a label (e.g., attack) to each word in the input text to identify the trigger word for an event that is pre-defined.
The second step is a multiple-class classification problem, which assigns a label to determine if the substring is an argument for the identified event in the first step.

\subsubsection{Formalization under ChatGPT}
Under ChatGPT, the event extraction task is formalized as a two-step text completion problem. \newline
{\bf Step 1}: We prompt ChatGPT to generate a text string to determine whether the input contains the trigger word with respect to a certain event type. 
If ChatGPT responds with a substring of the input, it denotes that the substring is the trigger with respect to the certain event. 
If ChatGPT generates \textit{null}, it indicates that the input does not contain an event with respect to the certain type. 
Suppose the number of pre-defined event types is $N$, the above process should be repeated $N$ times for a single input.  
For example, given the input text "\textit{On Sunday, a protester attacked an officer with a paper cutter.}",  the event type \textit{Attack}, the prompt to ChatGPT is: 
\begin{quote}
\textit{Given the sentence 'On Sunday, a protester stabbed an officer with a paper cutter', what is the trigger word of the attack event?" }.
\end{quote}
ChatGPT should generate "\textit{stabbed}", which denotes that the sentence contains an attack event and its event trigger is "\textit{stabbed}".\newline 
{\bf Step 2}: We use ChatGPT to generate a text string, which is an argument with respect to a certain role for the identified event in Step 1. 
Suppose that there are $M$ types of arguments for the event, we should repeat the prompting process $M$ times for one input.  If ChatGPT generates a substring from the input, it denotes that the substring is the argument with respect to a certain role for the event. 
If ChatGPT responds with \textit{null}, it denotes that the input text does not contain an argument with respect to the certain role for the event. 
In the example above, we have already identified an \textit{attack} event in the input and the trigger for the event is "\textit{stabbed}". In this step, we would like to extract the argument with the \textit{target} role for the \textit{attack} event.
We feed the following prompt to ChatGPT:
\begin{quote}
\textit{INPUT: On Sunday, a protester \#\#stabbed\@\@ an officer with a paper cutter. \newline 
The INPUT contains an "attack" event, and the "stabbed" is the event trigger (marked with \#\#\@\@ in the INPUT). \newline 
What was the target of the stabbing in the attack?"}
\end{quote}
In this case, ChatGPT should generate "\textit{an officer}", which represents that the target for the attack event is "\textit{an officer}".
Examples of the task formalization are shown in Figure~\ref{fig:task_formalization}.

\subsubsection{Datasets and Results}
We conduct experiments on English ACE2005 benchmark, and more details can be found in Appendix~\ref{appendix:dataset_event}.
Experimental results are shown in Table~\ref{exp:event}. The task is evaluated in terms of both the trigger extraction and the argument extraction.

We find that retrieving $k$NN with the SimCSE model as demonstrations outperforms 
the randomly selected strategy, achieving large performance boosts on the trigger extraction subtask (+5.2) and the argument extraction task (+4.8). 
This demonstrates that test-similar demonstrations can help ChatGPT capture feature information and thus improve performances. 
Besides, we also observe that the fine-tuned model as the demonstration retriever outperforms the SimCSE model. 

We find that the multiple-prompt strategy introduces a significant performance enhancement on the two event extraction subtasks, +1.8 on the trigger extraction subtask and +1.2 on the argument extraction. 
This phenomenon indicates that more supervision, which comes from the enlarged demonstration set in the multiple-prompt strategy, is helpful for the event extraction task.
Minor gains are observed on ACE2005 introduced by the reasoning strategy.
For the self-verification strategy, 
we find that it introduces minor performance boost in the argument extraction subtask, but no change in the trigger extraction subtask.
This phenomenon indicates that verifying ChatGPT's predictions in a new turn is helpful for the argument extraction task to correct a small fraction of prediction errors, while the strategy is not beneficial to the trigger extraction subtask.

In general,
after applying the proposed strategies, we are able to surpass the supervised baseline, filling the big gap between vanilla ChatGPT and the supervised baseline. 

\begin{table}[t]
\center
\small
\scalebox{0.9}{
\begin{tabular}{lcc}\toprule 
 & {\bf Peen WSJ} &  {\bf Tweets}  \\
{\bf Models} & {\bf (ACC)} &  {\bf (ACC)}  \\\midrule
RoBERTa-Large & {\bf 98.9} & 92.3  \\\midrule
\textit{\bf ChatGPT}~\textit{(few-shot)} &  &    \\
+Random demo  & 90.3& 84.7 \\
+SimCSE $k$NN  & 93.4& 88.2  \\
+FT $k$NN  & 98.2& 92.4 \\
+FT $k$NN+Multi & 98.7& 92.6 \\
+FT $k$NN+Multi+Reason & 98.7& 92.6 \\
+FT $k$NN+Multi+Reason+SV & {\bf 98.9} & {\bf 92.7} \\
\bottomrule
\end{tabular}
}
\caption{Experimental results on the part-of-speech datasets. We abbreviate the self-verification strategy as SV.}
\label{exp:pos}
\end{table} 

\subsection{Part-of-speech Tagging}
\subsubsection{Task Description}
Part-of-speech (POS) tagging~\citep{Brill1992ASR, Ratnaparkhi1996AME, Brants2000TnTA, Toutanvoa2000EnrichingTK, Gimpel2010PartofSpeechTF, Owoputi2013ImprovedPT, Santos2014LearningCR, Wang2015PartofSpeechTW, Yasunaga2017RobustMP, Chiche2022PartOS} is a task that aims to assign a part-of-speech label to each word in the given sequence based on its morphology (e.g., past tense), semantic meaning (e.g., move or action), and syntactic functions (e.g., preposition).
The POS task is normally formalized as a sequence labeling problem. Models assign a POS label (e.g., NN, NNS, NNP) to each word in the given sequence. 

\subsubsection{Formalization under ChatGPT}

Under ChatGPT, the POS task is formalized as a text completion problem, where ChatGPT is prompted to generate a POS-indicative  text for an annotated word in the sentence at a time. 
Specifically, we prompt ChatGPT to generate the POS for the marked word in the sentence with all POS options given. 
Suppose there are $N$ words in the sentence, the above prompting process should be repeated $N$ times. 

For example, given the input sentence "\textit{Vinken,61 years old.}",  and the marked word is "\textit{Vinken}".
We feed the following prompt to ChatGPT: 
\begin{quote} 
\textit{Part-of-speech categories are as follows: \newline} 
\textit{1. CC    Coordinating conjunction\newline} \textit{... \newline}
\textit{ 15. NNPS    Proper noun, plural\newline} \textit{...\newline} 
\textit{45. DT    Determiner.}\newline
{\it INPUT: \#\#Vinken\@\@, 61 years old  }\newline
{\it QUESTION: What is the POS tag for the word 'Vinken' which is marked by  \#\#\@\@ in the INPUT?}" 
\end{quote}
ChatGPT should generate "\textit{15. NNPS    Proper noun, plural}", denoting that the POS for "\textit{Vinken}" is NNPS.
Examples of the task formalization are shown in Figure~\ref{fig:task_formalization}.

\subsubsection{Datasets and Results}
We conduct experiments on two English benchmarks: WSJ Treebank and Tweets dataset. More details about the benchmarks are shown in Appendix~\ref{appendix:dataset_pos}.
We use accuracy for evaluation. 
Experimental results are shown in Table~\ref{exp:pos}. 

We observe that applying the SimCSE to retrieve $k$NN outperforms the random retriever, and there is a significant performance boost on both datasets, +3.1 and +3.5 on WSJ and Tweets, respectively.  
Additionally, we observe that using the fine-tuned model to retrieve $k$NN introduces a huge performance enhancement compared to the SimCSE model. 

The multiple-prompt strategy introduces a consistent performance improvement across the part-of-speech datasets.
This phenomenon indicates that the multiple-prompt strategy is benefit to the POS task by enlarging the demonstration set and providing more supervision signals. 
We also observe that the reasoning strategy does not introduce significant  performance boost on the POS datasets.
The explanation is that POS is a task that focuses more on superficial semantics, and does not require a deeper reasoning capacity.
Therefore, a reasoning model not necessarily can introduce a significant performance boost. 

By adopting the proposed strategies,  
 ChatGPT surpasses the supervised baseline, +0.06, +0.4 on the WSJ and Tweets datasets respectively. 


\begin{figure}
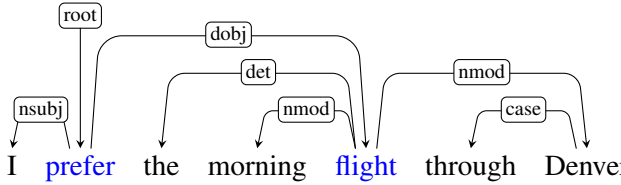

  \begin{minipage}[t]{0.3\linewidth}
    \centering
    \begin{dependency}
    \begin{deptext}[column sep=0.2cm]
   I \& {\color{blue}prefer} \& the \& morning \& {\color{blue}flight} \& through \&  Denver \& . \\
    \end{deptext}
    \depedge{2}{1}{nsubj}
    \depedge{5}{3}{det}
    \depedge{5}{4}{nmod}
    \depedge{5}{7}{nmod}
    \depedge{7}{6}{case}
    \deproot{2}{root}
    \depedge{2}{5}{dobj}
    \end{dependency}
  \end{minipage}%
  \caption{One dependency tree for sentence ``I prefer the morning flight to Denver''. In the sentence, {\color{blue} flight} is dependent to {\color{blue}prefer} with the dependency relation \textit{dobj}.}
\label{fig:dep_demo}
\end{figure}
\begin{table}[t]
\centering
\small
\scalebox{0.9}{
\begin{tabular}{lcc}
\toprule
 &  \multicolumn{2}{c}{\bf PTB}\\
{\bf Model}&  {\bf  (UAS)} &{\bf  (LAS)}  \\
\midrule
RoBERTa-Large & {\bf 96.87} & {\bf 95.34}  \\\midrule
\textit{\bf ChatGPT}~\textit{(few-shot)} &  &     \\
+Random demo  & 79.04 & 77.32   \\
+SimCSE $k$NN & 85.45 & 84.01  \\
+FT $k$NN  & 92.45 & 90.98  \\
+FT $k$NN+Multi& 93.72 & 92.20 \\
+FT $k$NN++Multi+Reason & 94.24  & 92.72   \\
+FT $k$NN+Multi+Reason+SV& 94.88 & 93.20  \\
\bottomrule
\end{tabular}
}
\caption{Experimental results for the dependency parsing task. We abbreviate the self-verification strategy as SV.}
\label{exp:dependency}
\vskip -0.15in
\end{table}

\subsection{Dependency Parsing}
\subsubsection{Task Description}
Dependency parsing~\citep{mcdonald2005online,dyer2015transition, Honnibal2015AnIN,  dozat2016deep, Kiperwasser2016SimpleAA, Zhang2017StackbasedMA,  Ma2018StackPointerNF, Dozat2018SimplerBM, mohammadshahi2019graph, yuan2019bidirectional, Qi2019UniversalDP,  gan2021dependency} is a task that aims to identify whether there are dependency relations between words in a sentence and determine the dependency relations. 
If  dependency relation holds between two words, the word that is relied on is called the head word, and the other is called the dependent word.
As shown in Figure~\ref{fig:dep_demo}, dependency parsing is to determine that "\textit{flight}" (dependent word) is dependent on "\textit{prefer}" (head word), and the dependency relationship between them is \textit{dobj} (direct object) for the given sentence "\textit{I prefer the morning flight to Denver.}".

Dependency parsing is usually implemented in two ways: transition-based methods and graph-based methods. The graph-based methods have become a more commonly used approach with the development of neural network.  Dependency parsing is usually formalized as a multi-class classification task in terms of deciding whether a relation between two words. 


\subsubsection{Formalization under ChatGPT}
Under ChatGPT, dependency parsing is formalized as a two-step text completion task.\newline
{\bf Step 1}: We use ChatGPT to rewrite the input sentence where dependent words for the given head word are marked with special tokens @\#, where @ denotes the start of a dependent word, and \# denotes the end of a dependent word. 
We use the example above as an illustration, where the input sentence is "\textit{I prefer the morning flight to Denver.}", the head word is "\textit{prefer}", and the prompt fed to ChatGPT is:
\begin{quote}
\textit{ The head word in the input is marked with @\#. Find the dependents of the head word. \newline}  
\textit{Input:I @prefer\# the morning flight to Denver}.
\end{quote}
ChatGPT should output:
\begin{quote} 
\textit{@I\# prefer the morning @flight\# to Denver.}"
\end{quote}
where "\textit{I}", "\textit{flight}" are marked and denote the dependent words for the head word "\textit{prefer}". If ChatGPT generates a sentence without the special token, it means that there are no dependent words for the given head word. 
Suppose that the given sentence is composed of $N$ words, we use one word at a time as the headword and will prompt ChatGPT $N$ times. \newline 
{\bf Step 2}: We use ChatGPT to generate yes or no to determine whether a given dependency relation holds between the head and the dependent word. 
As shown in Step 1,  "\textit{prefer}" and "\textit{I}" have a dependency relation. In this step, we will determine the category of the relation between the head word "\textit{prefer}" and the dependent word "\textit{I}". Suppose that there are $M$ types of dependency relations, we should prompt ChatGPT $M$ times, each of which corresponds to one type.
For example,
if we need to identify whether the dependency relation between "\textit{prefer}" and "\textit{I}" is \textit{direct object}, the prompt fed to ChatGPT is:
\begin{quote}
 \textit{@I\# @prefer\# the morning flight to Denver. whether the relation between "prefer" and "I" is the direct object?} 
\end{quote}
ChatGPT should generate "\textit{No}" in this case. 
Assuming that we obtain $C$ dependency word pairs from step 1, we should ask ChatGPT $M*C$ times for the given sentence. 
Examples of the task formalization under ChatGPT are shown in Figure~\ref{fig:task_formalization}.

\subsubsection{Datasets and Results}
We conduct experiments on the English Penn Treebank v3.0~\citep{marcus1993building} benchmark.
More details of the dataset can be found in Appendix~\ref{appendix:dataset_dep}. 
We use the unlabeled attachment score (UAS) and labeled attachment score (LAS) as evaluation metrics. 
Punctuations are ignored in all datasets during evaluation.
Experimental results are shown in Table~\ref{exp:dependency}. 

Results are shown in Table~\ref{exp:dependency}.
As expected, 
the SimCSE retriever  outperforms the random retriever, and the FT retriever outperforms the SimCSE retriever.  
ChatGPT with the multiple-prompt strategy obtains +1.3 improvement in terms of the UAS score. 
We also notice that the reasoning strategy introduces a performance boost on the WSJ dataset, +0.5 and +0.5 in terms of the UAS and LAS scores, respectively. 
For the self-verification strategy, it yields significant performance improvements on the WSJ dataset.
Overall, with the proposed strategies, we are able to bridge the gap between ChatGPT and  the supervised baseline, making ChatGPT's performance comparable to Roberta. 

\begin{figure}
\centering
{\linespread{0.5}\selectfont
The stock has been \underline{beaten} down for two days.\par
{\small \begin{flushleft} \hspace{1em} [\hspace{1.1em} A1\hspace{1.1em}]\hspace{5.0em}[beat.02]\hspace{0.1em}[\hspace{0.3em}A2\hspace{0.3em}]\hspace{0.3em}[\hspace{1.7em}TMP\hspace{1.7em}] \end{flushleft}}}

\vspace{4pt}
\begin{tabular}{|l|l|l|} 
\hline
sense id            & \multicolumn{2}{l|}{beat.02}             \\ 
\hline
sense                  & \multicolumn{2}{l|}{push, cause motion}  \\ 
\hline
 & A0 & causer of motion                  \\ 
\cline{2-3}
roles                       & A1 & thing moving                      \\ 
\cline{2-3}
                       & A2 & direction, destination            \\
\hline
\end{tabular}
\caption{An example of SRL. A0, A1 and A2 are {\it semantic roles} for the sense id ``beat.02''. The meanings of A0, A1 and A2 are respectively ``causer of motion'', ``thing moving'' and ``direction, destination''. The figure is adapted from the Figure 1 in the work of \citet{wang2021mrc}. }
\label{fig:srl_demo}
\end{figure}
\par
\begin{table*}[t]
\centering
\small
\hspace{-1cm}
\scalebox{0.9}{
\begin{minipage}{\textwidth}
\begin{tabular}{lcccc}
\toprule
&   \multicolumn{2}{c}{\bf  CoNLL 2009} &  \multicolumn{1}{c}{\bf  CoNLL 2005} & \multicolumn{1}{c}{\bf CoNLL 2012} \\
&  \multicolumn{1}{c}{\bf Predicate Disambiguation}& \multicolumn{1}{c}{\bf Argument Labeling} & \multicolumn{1}{c}{\bf Argument Labeling} & \multicolumn{1}{c}{\bf Argument Labeling}  \\
{\bf Model} &  \multicolumn{1}{c}{\bf (ACC)}& \multicolumn{1}{c}{\bf (F1)} & \multicolumn{1}{c}{\bf (F1)} & \multicolumn{1}{c}{\bf (F1)}  \\
\midrule

RoBERTa-Large & 97.3 & 93.3  &  89.3  &  87.6  \\\midrule
\textit{\bf ChatGPT}~\textit{(few-shot)} &  &  &  &   \\
+Random demo  & 83.2  &79.0 & 76.8  &  76.4  \\
+SimCSE $k$NN & 89.4   & 84.8   & 83.1  &  82.8   \\
+FT $k$NN  & 97.4  &  93.5 &   88.9  &  87.4 \\
+FT $k$NN+Multi & 97.8  &  93.8 &   89.8  &  88.2 \\
+FT $k$NN+Multi+Reason & 97.8  &  94.0 & 90.4  &  88.4  \\
+FT $k$NN+Multi+Reason+SV & {\bf 97.7} &  {\bf 94.1} &  {\bf 90.8}  &  {\bf 88.6} \\\bottomrule
\end{tabular}
\end{minipage}
}
\caption{Experimental results for the semantic role labeling task. We abbreviate the self-verification strategy as SV.}
\label{exp:srl}
\vskip -0.15in
\end{table*}

\subsection{Semantic role labeling}
\subsubsection{Task Description}
Semantic role labeling (SRL)~\citep{Zhou2015EndtoendLO, FitzGerald2015SemanticRL, Roth2016NeuralSR, He2017DeepSR, Tan2017DeepSR, Marcheggiani2017EncodingSW, He2018JointlyPP, Ouchi2018ASS, Zhang2021SemanticRL, Jia2022SpanBasedSR} is a task that aims to identify arguments for each predicate in a given sentence, along with determining semantic roles to the identified arguments. 
As shown in Figure~\ref{fig:srl_demo}, an SRL model identifies that "\textit{The stock}", "\textit{down}", "\textit{for two days}" are arguments for the predicate "\textit{beaten}", and semantic roles for the identified arguments are \textit{A1 thing moving}, \textit{A2 direction, desitnation}, \textit{TMP time, period or direction}, respectively.

SRL is normally formalized as a two-stage problem which involves: (1) word sense disambiguation for the predicate; 
and 
(2) semantic role labeling for arguments. 
The first stage is normally formalized as a multi-class classification task, which assigns a word sense label to the predicate in the given sentence. 
The second stage is solved under the sequence labeling framework, which assigns a label (e.g., B-ARG0, O) to each unit in the given sequence. If the label is B-ARG0, it denotes that the token is the beginning of an argument and it contains \textit{ARG0} semantic role for the predicate. 

\subsubsection{Formalization under ChatGPT}
Under ChatGPT, the SRL task is formalized as a two-step text completion problem.\newline 
{\bf Step 1}: We use ChatGPT to determine the word sense of the predicate.  
Use the example above as an illustration, where the sentence is "\textit{The stock has been beaten down for two days}", the predicate is "\textit{beaten}", there are three sense candidates for the predicate: (1)\textit{(Cause) pulsating motion that often makes sound}, (2)\textit{push, cause motion}; and (3)\textit{win over some competitor}.
We iteratively ask ChatGPT whether the predicate belongs to each of the three senses. \newline 
{\bf Step 2}: we use ChatGPT to  output an argument  that belongs to a certain semantic role with respect to the given predicate. Arguments are substrings of the input sentence. Suppose that there are $N$ semantic roles, we need to ask ChatGPT $N$ times, each of which corresponds to each role. 
Suppose that we would like to find the argument with the \textit{thing moving} semantic role, the input to ChatGPT is:
\begin{quote}  
\textit{What are the arguments  representing the meaning of 'thing moving'?}.
\end{quote}
ChatGPT should output "\textit{The stock}". If ChatGPT returns \textit{null}, it indicates that there is no argument in the sentence with the semantic role. 
Examples of the task formalization under ChatGPT are shown in Figure~\ref{fig:task_formalization}.

\subsubsection{Datasets and Results}
We conduct experiments on three English benchmarks: CoNLL2005~\citep{Carreras2005IntroductionTT}, CoNLL2009~\citep{Hajic2009TheCS} and CoNLL2012~\citep{Pradhan2013TowardsRL}. 
As for the predicate disambiguation subtask, we use accuracy as the evaluation metric. 
And for argument labeling, we report micro F1.

As shown in Table~\ref{exp:srl}, 
using a fine-tuned model to retrieve $k$NN yields better results than the SimCSE and random retriever.  
indicating that the FT model can learn more SRL task-related features than the SimCSE, leading to higher-quality retrieved demonstrations. 
The multiple-prompt strategy consistently improves performances in the semantic role labeling sub-tasks by +0.4, +0.3 with respect to the predicate disambiguation and argument labeling. 
 A minor improvement  is observed for  both the reasoning strategy and  the self-verification strategy.   
This phenomenon shows that verifying ChatGPT's output can help correct prediction errors caused by the randomness of decoding, but it has little effect on the predicate disambiguation subtask.


\section{Related Work}
\subsection{Large language models (LLMs)}
Large language models are models that aim to learn general language patterns and linguistic features by training in an unsupervised manner on large unannotated corpora~\citep{zhu2015aligning, 2019t5, lo2019s2orc, gao2020pile, Kopf2023OpenAssistantC}.
With the scale increases, LLMs achieve great performance boosts on various NLP tasks while unlocking emergent capabilities~\citep{xie2021explanation, wei2022emergent}. Other efforts~\citep{Sanh2021MultitaskPT, Wang2022SelfInstructAL, Longpre2023TheFC, Zhang2023ChineseOI} use human-instructions to boost LLM's ability.
Based on model architectures, LLMs can be categorized into three branches: 
\textbf{(1) encoder-only models}~\citep{devlin2018bert, liu2019roberta, sun2020ernie, clark2020electra, feng2020codebert, joshi2020spanbert, sun2020ernie, sun2021chinesebert} like BERT~\citep{devlin2018bert} are discriminative models that use a transformer~\citep{vaswani2017attention} encoder for getting the representation of a given sequence;
\textbf{(2) decoder-only models}~\citep{gpt1, dai2019transformer, keskar2019ctrl, radford2019language, brown2020language, chowdhery2022palm, ouyang2022training, zhang2022opt, scao2022bloom, zeng2022glm, touvron2023llama,  taori2023alpaca, chiang2023vicuna, peng2023instruction, anand2023gpt4all, OpenAI2023GPT4TR} like GPT~\citep{gpt1} are generative models that use the decoder of an auto-regressive transformer~\citep{vaswani2017attention} for predicting the next token in a sequence; 
\textbf{(3) encoder-decoder models}~\citep{lewis2019bart, raffel2020exploring, xue2020mt5} like T5~\citep{raffel2020exploring} are generative models that use both the encoder and decoder of the transformer~\citep{vaswani2017attention} model. Models finish downstream tasks by generating new sentences depending on a given input. 

\subsection{Adapting LLMs to NLP tasks}
In-context Learning (ICL)
has been adopted as a general strategy to apply LLMs to downstream NLP tasks. 
\citet{brown2020language} prompted LLMs to 
 generate textual responses (i.e., label words) conditioning on the given prompt with a few annotated examples without gradient updates.  
There are a variety of strategies to improve ICL performances on NLP tasks: \citet{li2021prefix, zhong2021factual, qin2021learning} propose to optimize prompts in the continuous space; \citet{rubin2021learning, das2021case, liu2021makes, gonen2022demystifying, su2022selective, wang2023gpt, wan2023gpt} investigate different strategies for selecting in-context examples; 
\citet{gonen2022demystifying} exploit different strategies for orders of in-context examples.
More advanced reasoning strategies 
\citet{wei2022chain, zhang2022automatic, han2021ptr, fu2022complexity, zhou2022least, sun2023text} 
also use in-context learning as the backbone.

\subsection{Generation Intermediate Rationales}
\citet{rajani2019explain} improve the interpretability of the model without sacrificing its performance by training a language model on the "explain-then-predict" commonsense answering dataset. 
Recently, \citet{nye2021show} find that a step-by-step computation “scratchpads” can improve LLM's performances on arithmetic, polynomial evaluation, and program evaluation tasks and etc.
\citet{wei2022chain} use manually annotated "chain-of-thoughts" prompts and greatly improve performances of LLM on complex reasoning tasks. 
After that, \citet{Li2022OnTA, fu2022complexity, Ye2022ComplementaryEF, Shao2023SyntheticPG} use higher reasoning complexity examples as demonstrations and further improve LLMs performances on complex reasoning tasks. 
\citet{Zhou2022LeasttoMostPE, Press2022MeasuringAN} decompose a complex problem into a series of simpler subproblems and then solve them step-by-step toward the final answer. 
\citet{zhang2022automatic, Kojima2022LargeLM, zelikman2022star, Chen2022ProgramOT, Sun2023EnhancingCP, Wang2023PlanandSolvePI, sun2023text} propose strategies to use LLMs generate explicit intermediate reasoning chains and then improve LLMs' complex reasoning ability with self-generated "chain-of-thoughts".

\section{Conclusion}
In this paper, we propose a collection of strategies in an attempt to push the limits of ChatGPT's performances, including
 (1) the one-input-multiple-prompts strategy; (2) the $k$NN as demonstrations strategy; (3) more natural formalizations for downstream tasks; (4) specific reasoning strategy across different tasks; (5) the verification strategy; (6) the paraphrase strategy. 
The proposed strategies 
efficiently address the factors that cause the subpar performance of ChatGPT: 
(1) token limit in the prompt does not allow for the full utilization of the supervised datasets;
(2) mismatch between the generation nature of ChatGPT and NLP tasks;
(3) intrinsic pitfalls of LLMs models, e.g., hallucination, overly focus on certain keywords, etc.
With the proposed strategies, ChatGPT achieves comparable or better results to the supervised RoBERTa on 17   of 21 datasets across 10 representative NLP tasks. 
\bibliography{custom}
\bibliographystyle{acl_natbib}

\appendix

\section{Evaluation Datasets}
\subsection{Question answering}
\label{appendix:dataset_qa}
\begin{itemize}
    \item {\bf SQuAD V2.0}: SQuAD V2.0\footnote{https://rajpurkar.github.io/SQuAD-explorer/} is a collection of 100K crowdsourced question-answer pairs which are originally from a set of Wikipedia articles.
    In this dataset, the answer to every question is a span of text from the context passage, or the question is unanswerable.
    \item {\bf TriviaQA}: TriviaQA\footnote{https://nlp.cs.washington.edu/triviaqa/} is a question-answering dataset which which includes 950K question-answer pairs from 662K documents collected from Wikipedia and the web. 
    In TriviaQA, all questions are answerable and answers may not be directly obtained from the given context. 
    \item {\bf MRQA OOD}: MRQA\footnote{https://mrqa.github.io/} out-of-domain (OOD) is a shared task which is to evaluate generalization to out-of-distribution data. 
    The test data contains 12 subsets, each from a held-out domain. 
\end{itemize}

\subsection{Commonsense Reasoning}
\label{appendix:dataset_commonsense}
\begin{itemize}
    \item {\bf CommonsenseQA}: CommonsenseQA\footnote{https://www.tau-nlp.sites.tau.ac.il/commonsenseqa} is a multiple-choice question answering dataset
    which requires commonsense knowledge to select one correct answer from four options to the question. The train/valid/test set contains 9,741, 1,221, and 1,140 questions, respectively.     
    \item {\bf StrategyQA}: StrategyQA\footnote{https://leaderboard.allenai.org/strategyqa/submissions/get-started} is an open-domain question answering dataset which requires implicit reasoning and logical inference to answer the question. There are 111 examples for the train set. 
    The dataset contains 2,821 examples in the train set and 490 examples in the test set.
\end{itemize}

\subsection{Natural language inference}
\label{appendix:dataset_nli}
\begin{itemize}
    \item {\bf RTE}: Recognizing Textual Entailment (RTE)\footnote{https://aclweb.org/aclwiki/Recognizing\_Textual\_Entailment} is an English dataset which is built from news and Wikipedia and from text entailment challenges. 
    \item {\bf CB}: CB\footnote{https://github.com/mcdm/CommitmentBank} is a corpus of 1,200 naturally occurring discourses whose final sentence contains a clause-embedding predicate under an entailment canceling operator (question, modal, negation, antecedent of conditional).
\end{itemize}

\subsection{Sentiment analysis}
\label{appendix:dataset_text}
\begin{itemize}
    \item {\bf SST-2}: SST-2 is a binary (i.e., positive, negative) sentiment classification dataset with 11,855 single sentences extracted from snippets of Rotten Tomatoes HTML files. 
    \item {\bf IMDB}: IMDB is a binary (i.e., positive, negative) sentiment classification dataset which includes 25,000 highly polar movie reviews for training and 25,000 for testing.
    \item {\bf Yelp}: Yelp is a binary (i.e., positive, negative) sentiment classification dataset that contains 560,000 highly polar Yelp reviews for training and 38,000 for testing.
\end{itemize}

\subsection{Named entity recognition}
\label{appendix:dataset_ner}
\begin{itemize}
    \item {\bf CoNLL 2003}: CoNLL 20033 is an English NER benchmark that includes four entity types: location, organization, person and miscellaneous. 
    We follow \citet{ma2016end} and use the same train/dev/test split. 
    \item {\bf OntoNotes 5.0}: OntoNotes 5.0 is an English NER dataset and contains 18 entity types. We use the standard train/dev/test split of CoNLL2012 shared task. 
\end{itemize}

\subsection{Entity-relation extraction}
\label{appendix:dataset_re}
\begin{itemize}
    \item {\bf ACE2004}: ACE2004\footnote{https://catalog.ldc.upenn.edu/LDC2005T09} is a multilingual information extraction benchmark. 
    The dataset contains 7 entity types and 7 relation categories. In this paper, we use English annotations and follow \citet{li2019entity} to split train/valid/test datasets.
    \item {\bf ACE2005}: ACE2005 is a multilingual training
corpus. It has six relation categories, and we process and split the dataset following the practice in \citet{li2019entity}
There are six subdomains in
the dataset: Broadcast Conversations (BC), Broadcast News (BN), Conversational Telephone Speech (CTS), Newswire (NW), Usenet Newsgroups (UN), and Weblogs (WL).
\end{itemize}

\subsection{Event Extraction}
\label{appendix:dataset_event}
\begin{itemize}
    \item {\bf ACE 2005}: ACE 2005 event corpus defines 8 event types and 33 subtypes, each event subtype corresponding to a set of argument roles.
There are 36 argument roles for all event subtypes.
In most of researches based on the ACE corpus, the 33 subtypes of events are often treated separately without further retrieving their hierarchical structures. The ACE 2005 corpus contains 599 annotated documents
and around 6000 labeled events, including English, Arabic and
Chinese events from different media sources like newswire
articles, broadcast news and etc. In this paper, use the English subset and follow \citet{liu2018event} to split the train/valid/test sets.
\end{itemize}

\subsection{Part-of-speech tagging}
\label{appendix:dataset_pos}
\begin{itemize}
    \item {\bf Penn WSJ}: Penn Treebank (PTB) is an English dataset corresponding to the articles of Wall Street Journal (WSJ). 
    In this paper, we use sections from 0 to 18 are used for training (38, 219 sentences, 912, 344 tokens), sections from 19 to 21 are used for validation (5,527 sentences, 131,768 tokens), and sections from 22 to 24 are used for testing (5,462 sentences, 129,654 tokens). 
\end{itemize}

\subsection{Dependency parsing}
\label{appendix:dataset_dep}
\begin{itemize}
    \item {\bf PTB}: PTB is an English dataset that contains 39,832 sentences for training and 2,416 sentences for testing. We follow \citet{Ma2018StackPointerNF} and use the same train/valid/test split. 
\end{itemize}

\subsection{Semantic role labeling}
\label{appendix:dataset_srl}
\begin{itemize}
    \item {\bf CoNLL2005}: CoNLL2005 contains a total number of 20 roles, while there are only 2.5 roles per predicate on average. we use sections 02-21 of WSJ corpus as Train data, section 24/23 as Dev/Test data, and three sections (CK01-03) of the Brown corpus as out-of-domain (OOD) data. 
    \item {\bf CoNLL2009}: CoNLL2009 builds on the CoNLL-2008 task and extends it to multiple languages. Data is provided for both statistical training and evaluation, which extract these labeled dependencies from manually annotated treebanks such as the Penn Treebank for English. 
    We follow \citet{wang2021mrc} to test on the English data and use the same train/valid/test split for evaluations.
\end{itemize}

\end{document}